\documentclass[letterpaper, 10pt, conference]{ieeeconf}

\IEEEoverridecommandlockouts 
\let\labelindent\relax

\usepackage[utf8]{inputenc}
\usepackage{amsfonts}
\usepackage[inline]{enumitem}
\usepackage{soul}
\usepackage{color}
\usepackage{tikz}
\usetikzlibrary{trees}
\usepackage[linesnumbered,ruled,vlined]{algorithm2e}
\usepackage{multirow}
\usepackage{multicol}
\usepackage{listings}
\usepackage{xurl}
\usepackage{hyperref}
\PassOptionsToPackage{hyphens}{url}\usepackage{hyperref}

\usepackage{braket}
\usepackage{subfiles}
\usepackage{booktabs,ragged2e}
\usepackage[flushleft]{threeparttable}

\usepackage{cite}

\usepackage{graphicx}
\graphicspath{{./Figures/}}
\usepackage[font=footnotesize]{subfig}
\usepackage{times}
\usepackage{amsmath}
\usepackage{amssymb}
\usepackage{bm}
\usepackage[font=small]{caption}
\usepackage{multicol}

\usepackage{titlesec}
\titlespacing\section{0pt}{3pt}{3pt}

\usepackage{xcolor, colortbl}

\definecolor{orange}{rgb}{1.0, 0.22, 0.0}

\title{Task-Oriented Grasping with Point Cloud Representation of Objects}

\author{Aditya Patankar$^1$, Khiem Phi$^2$, Dasharadhan Mahalingam$^1$, Nilanjan Chakraborty$^1$, and IV Ramakrishnan$^2$
\date{}
\thanks{$^{1}$The authors are with the Department of Mechanical Engineering, 
        Stony Brook University, USA.
        {\tt\small \{aditya.patankar, dasharadhan.mahalingam, nilanjan.chakraborty\}@stonybrook.edu.}}%
        \thanks{$^{2}$The authors are with the Department of Computer Science, 
        Stony Brook University, USA.
        {\tt\small \{kphi, ram\}@cs.stonybrook.edu.}}%
}

\begin{document}
\maketitle

\begin{abstract}
 In this paper, we study the problem of task-oriented grasp synthesis from partial point cloud data using an eye-in-hand camera configuration. In task-oriented grasp synthesis, a grasp has to be selected so that the object is not lost during manipulation, and it is also ensured that adequate force/moment can be applied to perform the task. We formalize the notion of a gross manipulation task as a constant screw motion (or a sequence of constant screw motions) to be applied to the object after grasping. Using this notion of task, and a corresponding grasp quality metric developed in our prior work, we use a neural network to approximate a function for predicting the grasp quality metric on a cuboid shape. We show that by using a bounding box obtained from the partial point cloud of an object, and the grasp quality metric mentioned above, we can generate a good grasping region on the bounding box that can be used to compute an antipodal grasp on the actual object. Our algorithm does not use any manually labeled data or grasping simulator, thus making it very efficient to implement and integrate with screw linear interpolation-based motion planners. We present simulation as well as experimental results that show the effectiveness of our approach. Website: 
 \url{https://irsl-sbu.github.io/Task-Oriented-Grasping-from-Point-Cloud-Representation/}
\end{abstract}


\section{Introduction}
\begin{figure}[!t]
    \centering
    {\includegraphics[width=0.5\textwidth]{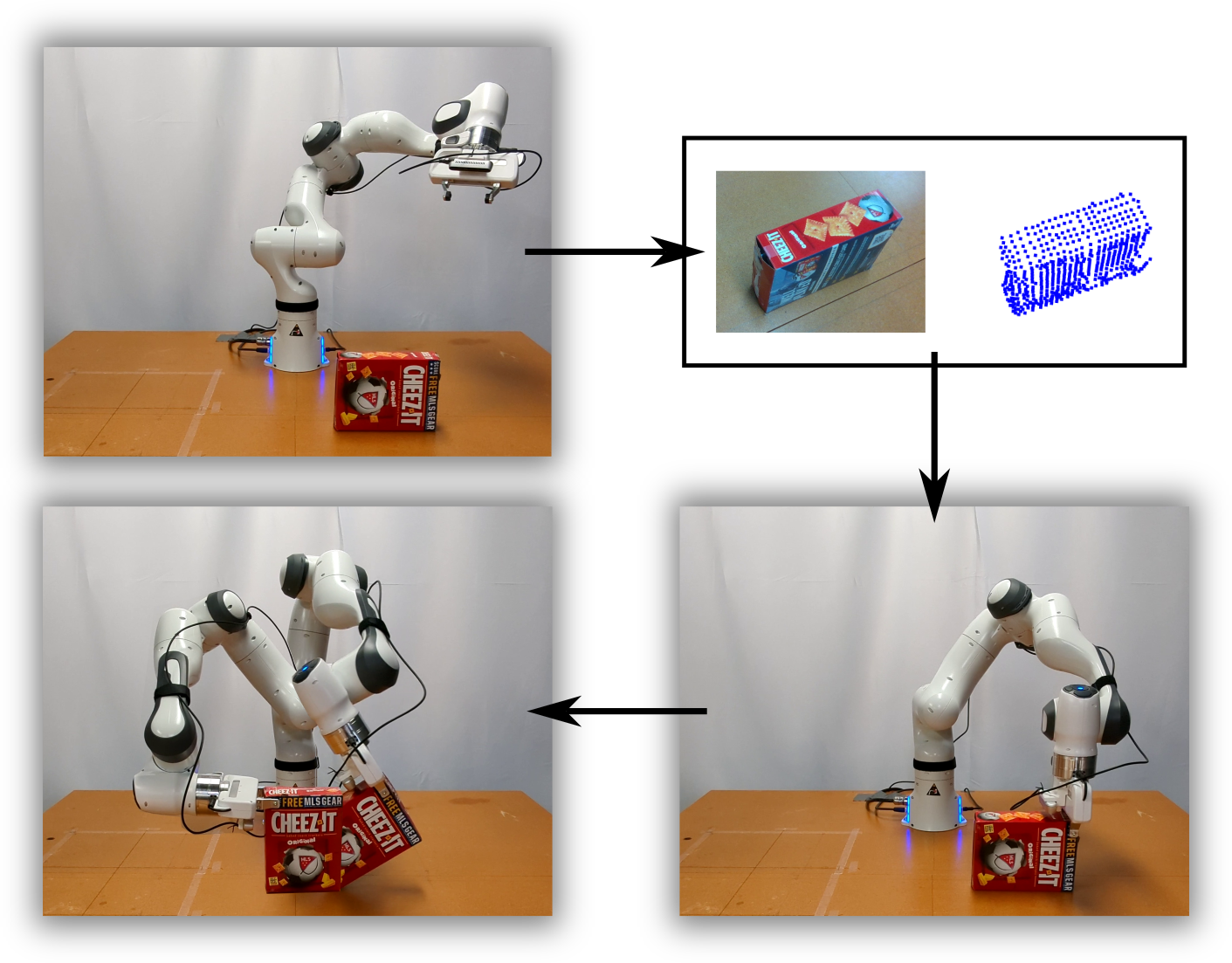}}
    \caption{Example task considered in this paper where the robot has to pivot the box to turn it upright. Using a eye-in-hand camera configuration (top left), the robot views the object as a partial point cloud (top right). Our algorithm enables the robot to select the appropriate grasping region (bottom right) to pivot the box about an edge to turn it upright (bottom left).}
\label{Fig:Pipeline}
\end{figure}

Grasp synthesis, in particular, sensor driven grasp synthesis is a fundamental problem in robotic manipulation. 
The extant literature on grasp synthesis has predominantly focused on pick and place tasks where the goal is to synthesize a grasp such that the grasp is not lost during movement (i.e., force closure conditions are satisfied). However, pick and place is just one type of manipulation task. In general, we grasp objects to give them a motion or apply a force (and/or moment) along (about) an axis. Figure~\ref{Fig:Pipeline} shows an example task of pivoting an object to give it a motion so as to set it upright. Other examples include grasping a screw driver to use it for fastening/unfastening a screw, grasping a spatula to use it, grasping a box of sugar to pour from it. Thus, task-oriented grasping is a key capability for robots to have to expand their usage beyond pick-and-place tasks. Furthermore, for a robot to operate in environments that are not specifically engineered for it, it has to sense the objects to be grasped and may not have {\em a priori} knowledge about the geometry of the objects (i.e., CAD models of the objects may not be available). {\em Therefore, the goal of this paper is to develop an algorithm for task-oriented grasp synthesis with partial point cloud information about the objects}. 



Extant literature on sensor-based task-oriented grasp synthesis has focused on affordance-based grasping~\cite{kokic2020learning, wen2022catgrasp}, where human intuition about good regions to grasp for a task label is used to create datasets, and deep learning-based techniques are used to generate a grasp. However, the motion (or force/moments) to be applied to the object after grasping, which is also key for selecting a grasp region for many tasks, is usually not considered. For example, for the pivoting task shown in Figure~\ref{Fig:Pipeline}, the region where the object should be grasped depends on the axis of pivot after grasping. Even if multiple grasps are provided as examples, the suitability of the grasp depends on the pivoting axis, and good grasps for one pivoting axis is a bad grasp for another axis. Similarly, in tasks like pouring from a symmetric object, the region of grasping depends on the axis about which the object should be rotated for pouring. 

Therefore, in this paper, we focus on using motion to formalize the notion of task and present an algorithm for grasp synthesis to generate a desired motion. 
Following~\cite{fakhari2021computing}, we define a task as a unit screw about which the robot should generate a desired motion after grasping. More generally, we can think of a task as a sequence of unit screws about which the fingers/manipulators need to generate a desired motion after grasping. Although there are several formalizations of a task in the literature on computing metrics for task-oriented grasping~\cite{Borst2004, pollard1994parallel, haschke2005task, krug2016analytic, song2020grasping}, we use the screw geometry-based approach, as it allows us to directly connect our grasp planning to motion planning after the grasp~\cite{mahalingam2022human,sarker2020screw}.

Given the gripper contact locations, and a task specified by a unit screw, one can evaluate the suitability of the locations for performing the task by computing a grasp quality metric using a convex optimization problem, namely a second order cone program~\cite{fakhari2021computing}. However, solving the inverse problem of computing the contact locations given the task is much more challenging, even with the knowledge of the object geometry, since technically the problem becomes a non-convex optimization problem~\cite{patankar2020hand}. Furthermore, the potential presence of environmental contacts during manipulation, as well as partial knowledge of the object geometry, say, in the form of a point cloud, makes the problem more challenging. Therefore, more precisely, we have to solve the following problem: {\em Given a partial point cloud representation of an object and a task defined by a unit screw, compute an antipodal grasp for manipulating the object, where manipulation may also require environmental contact with the object}.

We present a novel grasp synthesis algorithm that uses a bounding box approximation of an object to generate grasps to provide a motion to the object after grasping. We assume that the object is being sensed with a RGBD camera, from which the partial point cloud is extracted. A key step in our approach is to learn a function that takes in a task screw and contact locations as input, and outputs a metric denoting the suitability of the contacts for manipulating a cuboid shape. We show that by using this function on the {\em simplest possible geometric representation of the partial point cloud, namely, a bounding box}, we can generate a good grasping region on the actual object. Experimental results are provided to validate our approach and show that it can be combined with a motion planner~\cite{mahalingam2022human,sarker2020screw} to generate a grasp for performing tasks like pouring and pivoting on different objects. Note that in pivoting there are contacts of the object with the environment apart from the gripper object contact, while in pouring there is no object environment contact. Our method can generate grasps for both of these categories of manipulation tasks. The neural network we learn uses data generated from the grasping force optimization algorithm~\cite{fakhari2021computing} on cuboids.  Therefore, we do not need CAD models of objects or example grasps from a dataset to generate the grasps. 

\section{Related Work}
\label{sec:rw}
Sensor driven grasp synthesis has been studied primarily for pick and place tasks, although recent efforts have also focused on task-oriented or task-relevant grasping. Since this paper is about task-oriented grasping we will limit our discussion to the task-oriented grasping literature. A more detailed discussion on grasp synthesis can be found in the excellent review article~\cite{newbury2022deep}.

There are two key aspects to task-oriented or task-relevant grasp synthesis, namely, (a) the knowledge of the regions of an object that can/cannot be grasped for performing a particular task, and (b) the motion (or forces/moments) to be applied to the object to accomplish the task. As stated before, the sensor-driven grasp synthesis literature~\cite{kokic2017affordance, detry2017task, do2018affordancenet, ardon2020self, deng20213d, fang2020learning} primarily focuses on the aspect (a) above. The notion of task-relevance is based on (usually) human provided labels in datasets~\cite{murali2021same, sun2021gater} associating a $6-$degree-of-freedom (DOF) gripper pose or contact regions on objects to a task label (verb). Such labels or task affordances are sufficient for some tasks like handover, or for moving objects that has a handle to hold it. However, there are other tasks like pivoting, pouring, turning a handle, etc., where the grasp selection also depends on the motion to be performed after grasping. Our paper focuses on these type of tasks where the motion to be performed after grasping is key to the successful accomplishment of the task. 

In~\cite{fang2020learning}, the authors present a deep-learning based algorithm for grasp selection as well as a policy for motion after the grasp selection for two types of tasks, namely, hammering and sweeping. The focus is on using the same object for these two types of tasks and it was assumed that the sensor data is a depth image data from a fixed overhead camera. This approach does not easily generalize to our scenarios where we have partial point cloud data from a eye-in-a-hand configuration. Further generalization to tasks like pivoting where the objects can be in contact with the ground during manipulation after grasping is also not clear. As such there are no approaches or datasets for task-oriented grasping that has data for performing tasks where there may be contacts other than finger object contact which constrains the motion during manipulation and has to be taken into consideration for grasp synthesis. 

In contrast, we take an alternate route for sensor-driven grasp synthesis, where we first formalize the notion of a task using the screw geometry of motion. Further, we use task-oriented grasp metrics to learn a function to evaluate the suitability of an antipodal grasping region for a given task specified by a unit screw. Grasp metrics like the Ferrari-Canny metric~\cite{ferrari1992planning} has been used for pick-and-place tasks~\cite{mahler2017dex, ten2017grasp, alliegro2022end, lou2021collision, lou2020learning, kasaei2021mvgrasp, mousavian20196} and our idea of using the task-oriented grasp metric for grasp synthesis was inspired from these works.

\section{Task-Oriented Grasping}

In this section we formalize the notion of a task and present our problem formulation of sensor-driven task-oriented grasping. Our formalization is based on the fact that, in the analysis of rigid body motion, a {\em screw} is the underlying geometric entity  for both twists (i.e., linear and angular velocity pairs) as well as wrenches (i.e., force/moment pairs).
%
%
%


We define a task as a unit screw about (along) which a desired motion has to be generated for an object by using a robotic hand that has to apply a wrench to generate the desired motion. Henceforth, we will refer to this unit screw as the \textit{task screw}, $\mathcal{S}$, which is a line in space. We represent $\mathcal{S}$ using its Plucker coordinates $(\bm{l}, \bm{m})$, where,  $\bm{l} \in \mathbb{R}^3$ is a unit vector along $\mathcal{S}$ and $\bm{m} = \bm{p} \times \bm{l}$ is the moment with $\bm{p} \in \mathbb{R}^3$ being any point on $\mathcal{S}$. We assume that the gripper to be used is a parallel jaw gripper and thus the grasp is antipodal.

{\bf Problem Statement}: {\em Given a point cloud representation, $\mathcal{O}$, of a single object and a task screw, $\mathcal{S}$, in terms of its Plucker coordinates $(\bm{l}, \bm{m})$, compute a grasping region for an antipodal grasp, $\mathcal{I}_O$ $\subseteq$ $\mathcal{O}$ such that the grippers after grasping the object can apply the desired motion on the object specified by the task screw $\mathcal{S}$. }

Note that both $\mathcal{O}$ and $\mathcal{S}$ are expressed with respect to a fixed object reference, $\{\mathbf{O}\}$. 
The ideal grasping region, $\mathcal{I}_O$, can be used to compute a set of $6$-DOF end-effector poses, $\mathcal{E}$, such that upon closing the parallel jaw gripper at any one of the poses we would be able to apply the necessary contact wrenches thereby imparting the desired motion on the object and accomplishing the task.

\section{Task Oriented Grasping From Point Clouds}

\begin{figure*}[t!]
    \centering
        \includegraphics[width=0.3\textwidth]{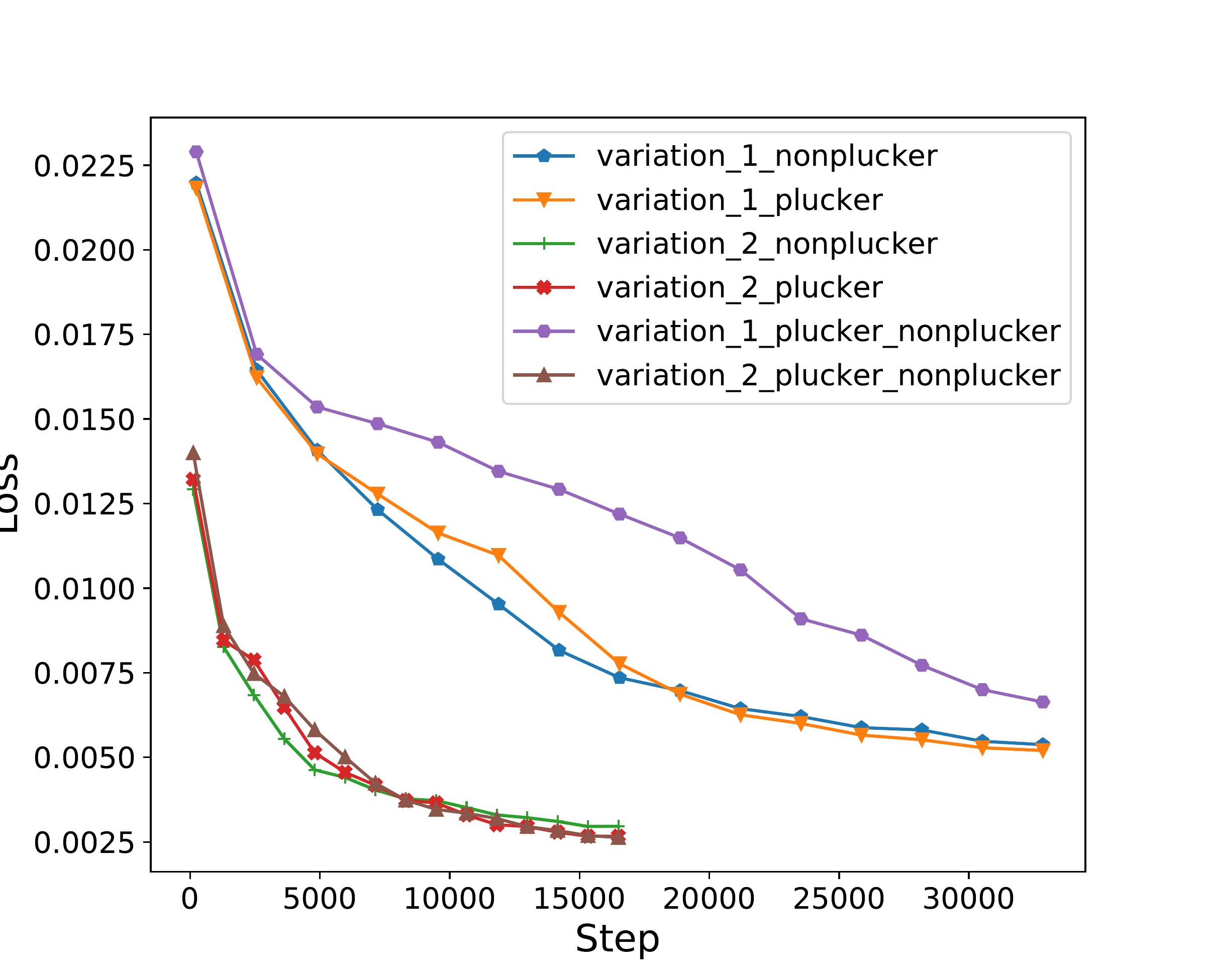}
        \includegraphics[width=0.3\textwidth]{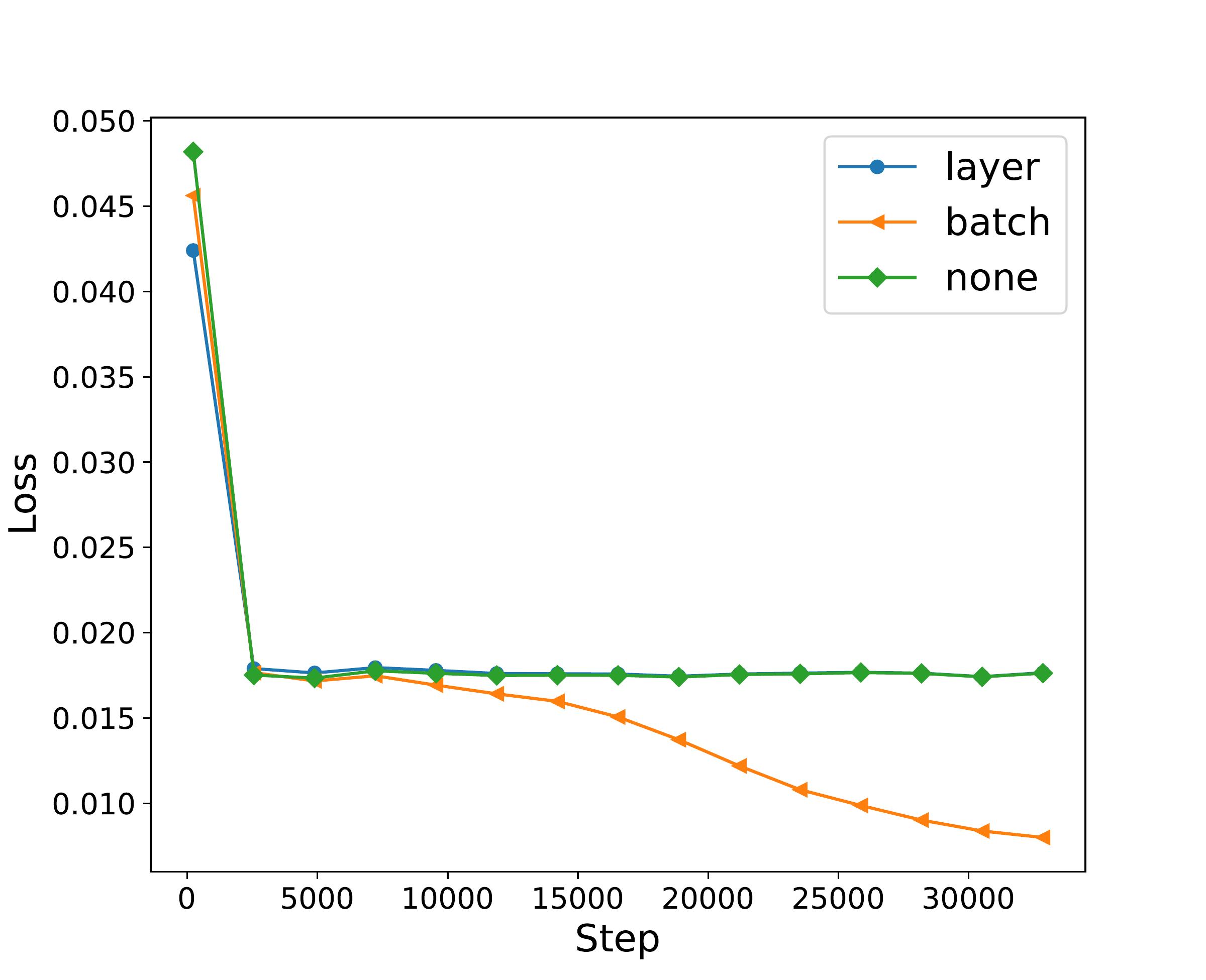}
        \includegraphics[width=0.3\textwidth]{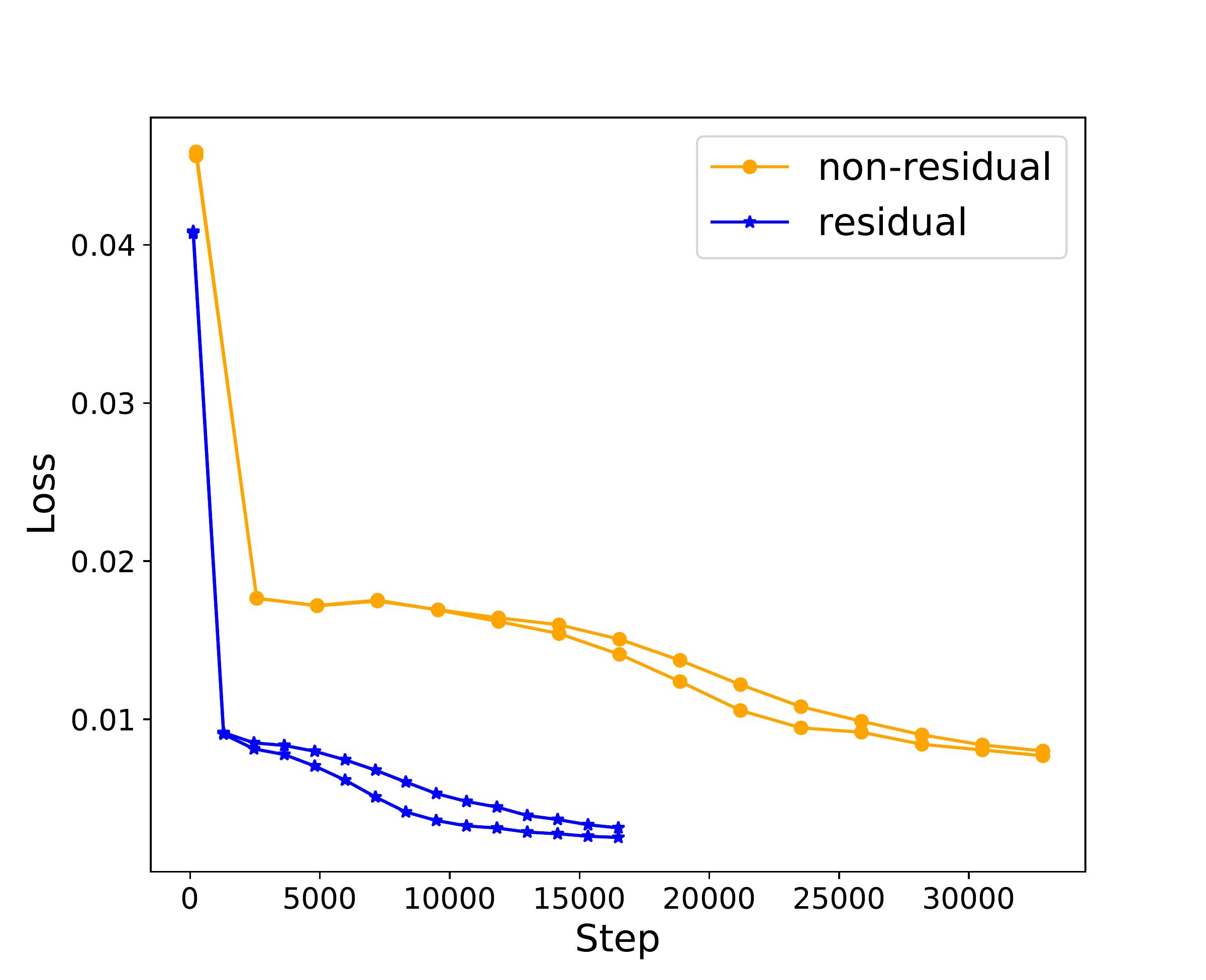} 
\caption{\textbf{\textsc{Multi-layer Perceptron (MLP) ablation studies:}} From left to right are the training loss across steps of models using different feature vectors, normalization strategies, and residual connections. We incorporate BatchNorm and skip connections in our final MLP design and utilize a 12-dimensional Plucker-coordinate-based vector.  }
\label{Fig:orderemb}
\end{figure*}

Our overall grasp computation strategy can be divided into the following steps: (1) Compute a bounding box $\mathcal{B}$ of the partial point cloud data, (2) Evaluate the grasp metric on the bounding box, $\mathcal{B}$, to (implicitly) find the ideal grasping region $\mathcal{I}_B$ on $\mathcal{B}$, (3) Convert $\mathcal{I}_B$ to an ideal grasping region $\mathcal{I}_O$ on the object surface. 
The computation of the bounding box of a point cloud data is a classical problem in computer vision and we use off-the-shelf algorithms for step $1$ (please see the results section for more details). The steps $2$, and $3$ are described below and forms the bulk of the technical contribution of this paper.

\begin{figure}[h]
    \centering
    {\includegraphics[width=0.2\textwidth]
    {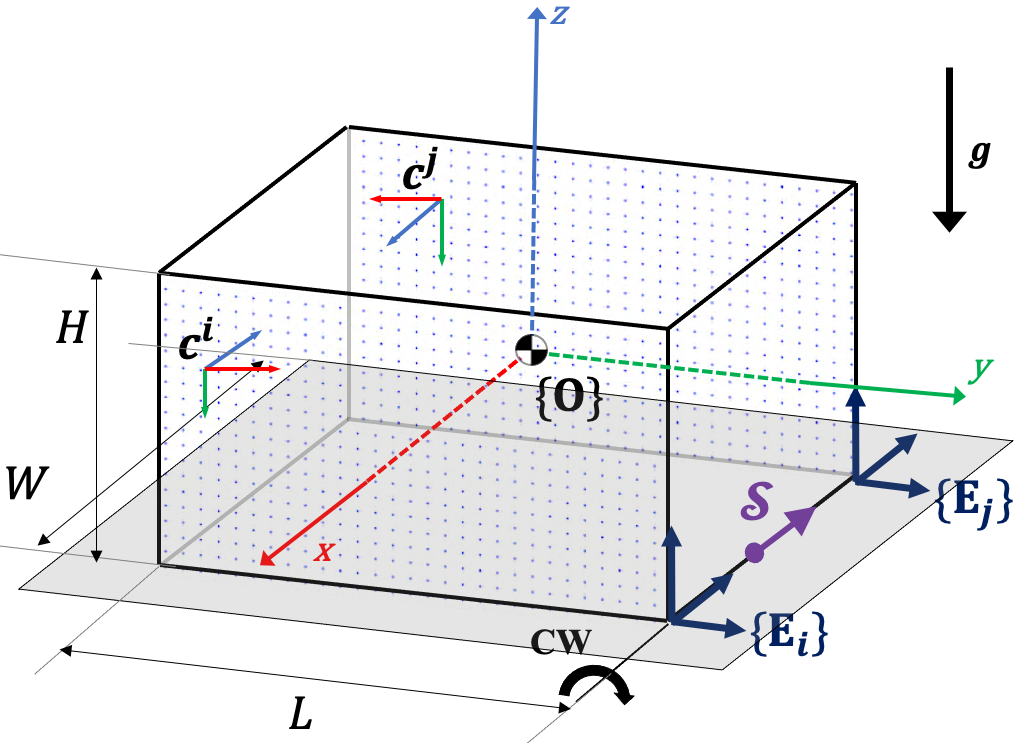}}
    \label{Fig:one}
    {\includegraphics[width=0.25\textwidth]{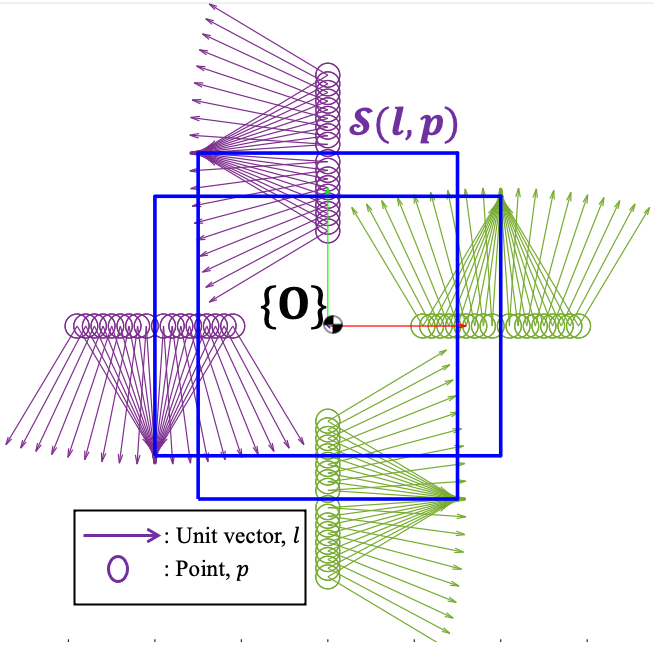}} 
    \label{Fig:two}
    \caption{(a) Pivoting a cuboid using antipodal contact locations, $\bm{c}^i$ and $\bm{c}^j$, sampled along its length $L$; (b) Variation in the screw parameters of a task screw $\mathcal{S}(\bm{l}, \bm{p})$ as captured in our dataset.} 
\label{Fig:Data_Generation}
\end{figure}

\subsection{Computing Ideal Grasping Region on Cuboids}
Since the bounding box of an object is a cuboid, we first present a solution for computing antipodal grasps on a cuboid object. To compute an antipodal grasp on a cuboid to give a constant screw motion about/along a given screw axis, a naive strategy is to discretize the faces of the cuboids, compute the grasp metric for antipodal grasps using the formulation in our previous work~\cite{fakhari2021computing}, and then select the grasp with the maximum value. However, the grasp metric computation for each antipodal grasp involves solution of a second order cone program (SOCP). Thus, the time taken for performing this computation would not allow real-time grasp computation. Therefore, we train a neural network that allows for a fast approximation of the grasp metric value for a given antipodal contact location and task screw.

\subsubsection{Approximating the Grasp Quality Metric}
The neural network to approximate the grasp quality metric approximates the target function $f$ : $\mathcal{X}$ $\rightarrow$ $\mathcal{Y}$. 
Each element, ${\bf x}$ $\in$ $\mathcal{X}$ $\subset \mathbb{R}^{12}$, is of the form: ${\bf x} = {[\bm{c}^i, \bm{c}^j, \bm{l}, \bm{m}]}^T$, where the vectors, $\bm{c}^i$ and $\bm{c}^j$, are a pair of antipodal contact locations sampled from the two opposite faces, $F^i$ and $F^j$, of the cuboid, $\mathcal{B}$. The last two elements are the Plucker coordinates $(\bm{l}, \bm{m})$ of the task screw, $\mathcal{S}$. The output of the neural network, $y$ $\in$ $\mathbb{R}$, is the value of the normalized task-dependent grasp metric, $\eta$. The output space $\mathcal{Y}$ can be defined as $\{y \in \mathbb{R}| 0 \leq y \leq 1\}$. 
All parameters of the input vector, $\bf{x}$, are expressed with respect to $\{\mathbf{O}\}$. A key aspect of learning the function $f$ is generating the training data, which we discuss next.

\subsubsection{Training Data Generation}

To generate training data, we select the task of pivoting a cuboid about one of its edges, in simulation, as shown in Fig.~\ref{Fig:Data_Generation}(a). The objective is to generate a constant screw motion about the task screw, $\mathcal{S}$, which lies along the pivoting edge. 
A key reason for selecting the pivoting task for data generation is that it includes object-environment contact, and by setting the environmental contact force to zero, we can consider situations with no object-environment contact. 
Each pair of antipodal contact locations, $\bm{c}^i$ and $\bm{c}^j$ along with the object-environment contacts at $\{\bf{E}_i\}$ and $\{\bf{E}_j\}$ and the task screw, $\mathcal{S} (\bm{l}, \bm{m})$, form one instance of the optimization formulation in~\cite{fakhari2021computing}. Upon solving the formulation for each grasp, ($\bm{c}^i$, $\bm{c}^j$), we have an associated metric value, $\eta$, which denotes the maximum magnitude of moment that can be generated about $\mathcal{S}$. 


The parameters that affect the value of $\eta$, in addition to the locations of the antipodal and object-environment contacts, are the weight of the object, the location of the task screw, $\mathcal{S}$, and the contact friction. The weight of the object is kept constant for all instances because varying the weight only changes the magnitude of $\eta$ and not the actual order of ($\bm{c}^i$, $\bm{c}^j$) imposed by $\eta$. For the friction coefficient at $\bm{c}^i$ and $\bm{c}^j$, we randomly sample $50$ values from a normal distribution, ${\mu_c}_{i,j} \sim \mathcal{N}(0.3, 0.05)$ and compute $\eta$ by solving the optimization formulation in~\cite{fakhari2021computing}. The final value of $\eta$ associated with ($\bm{c}^i$, $\bm{c}^j$) is the average of these $50$ values. For each grasp, the coefficient of friction is kept the same for both object-robot contacts, i.e. ${\mu_c}_{i} = {\mu_c}_{j}$. The coefficient of friction at both the object-environment contacts is kept constant, ${\mu_E}_{i,j}=0.4$. 








To capture the variation in the task screw, $\mathcal{S}$, and its effect on the value of $\eta$, we vary the length, $L$, of one of the two faces, $F^i$ or $F^j$, by a positive scalar $\delta$. This allows us to generate different cuboids where the length, $L$, of one face is different from the length, $L \pm \delta$, of its parallel face. The variation in the unit vector, $\bm{l}$, and the point, $\bm{p}$,  corresponding to the task screw, $\mathcal{S}$, in our dataset is shown Fig.\ref{Fig:Data_Generation}. Each unit vector, $\bm{l}$, and point, $\bm{p}$, correspond to the pivoting edge of a cuboid where the lengths of the parallel faces have been varied and $\bm{p}$ lies at the center of the pivoting edge. In our dataset we have a total $144$ such cuboids. We sample antipodal contacts, $\bm{c}^i$ and $\bm{c}^j$ along min$(L,L \pm \delta)$, while keeping the width and height constant for all the cuboids. 
The metric values $\eta$ are normalized for each cuboid separately before being used for training. To summarize, for each cuboid, we sample $(\bm{c}^i, \bm{c}^j)$ along the faces perpendicular to the task screw, $\mathcal{S}$, and solve multiple instances of the optimization formulation~\cite{fakhari2021computing}, thereby computing the normalized metric value, $y$, associated with each datapoint, $\bf{x}_n$. For generating the dataset the optimization formulation has been implemented using the CVX toolbox\cite{grant2014cvx} in MATLAB with the SDPT3 solver.








\subsubsection{Feature Vector Design}

We will now discuss the different variations of feature vector input and datapoints used to train our neural network architecture. Using different representations of the task screw, $\mathcal{S}$, along with $(\bm{c}^i, \bm{c}^j)$, we generate $3$ different input feature vectors with Plucker coordinates, $(\bm{l}, \bm{m})$, non-Plucker coordinates, $(\bm{l}, \bm{p})$ and a combination of both, $(\bm{l}, \bm{m}, \bm{p})$. Note that the $1^{\text{st}}$ and $2^{\text{nd}}$ feature vectors are $12$ dimensional and the $3^{\text{rd}}$ is $15$ dimensional. The antipodal contacts, $(\bm{c}^i, \bm{c}^j)$, and the location of the task screw, $\mathcal{S}$, influence the value of the task-dependent grasp metric $\eta$. Therefore, as additional features, we include the magnitude of the moment arms between $(\bm{c}^i, \bm{c}^j)$-$\{\mathbf{O}\}$, $\{\mathbf{O}\}$-$\bm{p}$ and $(\bm{c}^i, \bm{c}^j)$-$\bm{p}$. The $3$ moment arms along with the antipodal contact locations $(\bm{c}^i, \bm{c}^j)$ and the task screw representation $(\bm{l}, \bm{m}, \bm{p})$ form our fourth and final input feature vector $\bf{x}_n$ $\in$ $\mathbb{R}^{18}$. We would like to highlight the fact that there is no change in the normalized metric values, $y$, for the different variations of the input feature vector, as it is calculated using our optimization formulation \cite{fakhari2021computing}. In terms of the number of datapoints used for training, we use two variations, as shown in Fig.~\ref{Fig:Data_Generation}. The first variation (shown in green) consists of 46,512 datapoints from 72 cuboids. For the second variation, we include additional datapoints (shown in purple) bringing the total to 93,043 datapoints and 144 cuboids.


\subsubsection{Neural Network Ablation Study}
To approximate our task-dependent grasp metric, we use a fully connected Multilayer Perceptron (MLP) implemented using PyTorch. We experimented with all the feature vectors described above as input and obtained the lowest training loss using an input feature vector with Plucker coordinates when trained with the first variation of our data. This is shown in the leftmost graph of Fig.~\ref{Fig:orderemb}. Thus, for our input, we choose a feature vector using Plucker coordinates. 
Increasing the number of hidden layers in the MLP beyond eight layers does not significantly reduce training loss. We also observed that the ReLU activation function significantly reduces the training loss. Additionally, we experimented with the usage of skip-connections. Skip connections were introduced by \cite{he2016deep} for convolutional neural networks (CNN) to gain accuracy with increasing depth. Our tests have shown that the addition of skip-connections also drastically lowers the training loss, and this is shown in the rightmost graph in Fig.2. To determine the type of normalization to use for our network, we experimented with BatchNorm \cite{ioffe2015batch} and LayerNorm \cite{ba2016layer}. We found that BatchNorm yielded significantly lower training loss compared to layer normalization or when normalization was not applied. This comparison is shown in the middle graph in Fig.~\ref{Fig:orderemb}. Based on the observations made above, our MLP comprises of eight hidden layers that uses the ReLU activation function, normalized by BatchNorm and with added skip-connections. Our network is trained using the Stochastic Gradient Descent algorithm and the Mean Squared Error (MSE) loss function on a desktop with NVIDIA RTX 3060. We set our learning rate to $0.001$ and train our model for $150$ epochs using a batch size of $150$. 

{\bf Computation of grasp metric on bounding box}: We use the neural network above, which is trained to approximate the grasp quality metric, with the bounding box $\mathcal{B}$ of the partial point cloud $\mathcal{O}$ of the object to form an implicit representation of the ideal grasping region $\mathcal{I}_B$. Algorithm~\ref{ideal-region-algorithm}, lines $2$-$4$ explains this. In line $2$ of Algorithm~\ref{ideal-region-algorithm} we generate a grid on two parallel faces, $F^i$ and $F^j$, of the bounding box $\mathcal{B}$ and select antipodal contact locations, $(\bm{c}^i, \bm{c}^j)$ from the same. The faces are selected according to the dimensions of the bounding box and maximum allowable width of the gripper opening, $g_w$. The output of our neural network provides an order for each pair of antipodal contacts, $(\bm{c}^i$, $\bm{c}^j)$, based on their corresponding metric value $y$, forming an implicit representation of the ideal grasping region $\mathcal{I}_B$ corresponding to the bounding box, $\mathcal{B}$.

\subsection{Computing the Ideal Grasping Region on Objects }

Our goal now is to use $\bf{y}$, to obtain an ordering of the actual points in $\mathcal{O}$ that is described by lines $6$-$8$ of Algorithm~\ref{ideal-region-algorithm}. The point cloud, $\mathcal{O}$, is projected on either of the two parallel faces, $F^i$ or $F^j$, of the bounding box $\mathcal{B}$. Let us denote the point cloud projected on the face $F^i$ as $P^i$. We create a grid, $G^i$, along $F^i$ using the corresponding contacts $\bm{c}^i$ as the vertices of each element. These elements are assigned a metric value $y_{\text{avg}}$, which is the average of the metric values corresponding to the contacts at the vertices of the element. Essentially, we have converted a vertex-based representation of approximated metric values along $F^i$, to an element-based one. Next, we check which points of $P^i$ lie in a particular element of $G^i$, and the metric value, $y_\text{avg}$, associated with that element is assigned to the points occupying it. Each point in $P^i$ and their corresponding metric value, $y_\text{avg}$, is projected back to its original location in $\mathcal{O}$. Using a threshold, $y_\text{th}$, we extract the points with $y_\text{avg} \geq y_\text{th}$ as explained in lines $9$-$11$ of Algorithm~\ref{ideal-region-algorithm}. These points would correspond to the ideal grasping region $\mathcal{I}_O$ $\subseteq$ $\mathcal{O}$. Thus, we now have a set of points, $\mathcal{I}_O$ $\subseteq$ $\mathcal{O}$, where if the object is grasped then the manipulator would be able to generate the necessary moment about the task screw, $\mathcal{S}$.



\begin{figure}[!t]
    \centering
    {\includegraphics[width=0.35\textwidth]{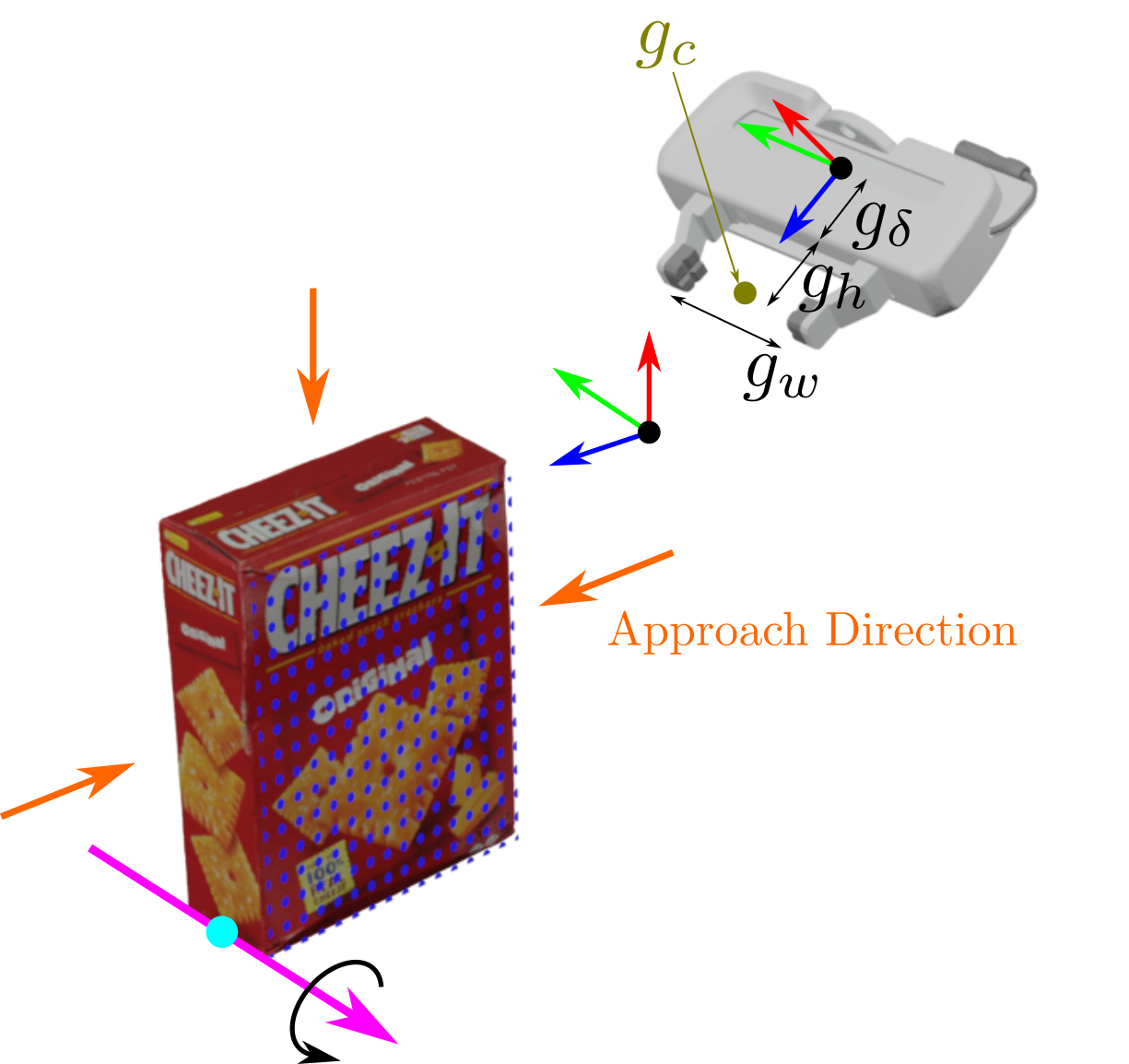}}
    \caption{The grasping pose of the end-effector is computed based on (a) the faces corresponding to the bounding box, $\mathcal{B}$, (b) the orientation of the reference frame attached at the wrist joint of the Franka Emika Panda.}
\label{Fig: Reference_Directions}
\end{figure}

\SetCommentSty{CommentFont}
\SetKwInput{KwInput}{Input}
\SetKwInput{KwOutput}{Output}
\begin{algorithm}[t!] 
\DontPrintSemicolon
  \KwInput{
  $\quad$ Point Cloud, $\mathcal{O}$, Screw Axis, $\mathcal{S} (\bm{l}, \bm{m})$ \\
  $\quad$ Gripper Geometry, $G$, Threshold value, $y_{\text{th}}$}
  \KwOutput{
  $\quad$ Ideal Grasping Region, $\mathcal{I}_O$ }
  
  $\mathcal{B} \gets  \textit{computeBoundingBox}\{\mathcal{O}\}$
  
  
  $\bm{c}^i, \bm{c}^j \gets \textit{sampleContacts\{$\mathcal{B}$, $F^i, F^j, G$\}}$
  
  ${\bf x}_n \gets {[ \bm{c}^i, \bm{c}^j, \bm{l}, \bm{m}]}^T$
  
  Concatenate all ${\bf x}_n$ into a single vector, $\bf{x} \in \mathbb{R}^{n \times 12}$ where $n$ is total number of antipodal contact locations.
  
  $\bf{y}$ $\gets$ $\textit{metricNeuralNet}\{ \bf{x} \}$, $\bf{y} \in \mathbb{R}^n$

  
  $\mathcal{P}_i$ $\gets$ $\textit{projectCloud}\{ \mathcal{B} \}$
  
  $\mathcal{G}^i$ $\gets$ $\textit{generateGrid}\{ \mathcal{B}, \mathcal{P}_i, F^i, \bf{y} \}$
  
  $(\mathcal{O}, \bf{y}_{\mathcal{O}}$) $\gets$ $\textit{assignMetricValue}\{ \mathcal{P}_i, \mathcal{G}^i, \bf{y}\}$
  
  \For{$y$ \text{in} $\bf{y}_{\mathcal{O}}$}
  {
  \If {$y \geq y_{\text{th}}$}
    { 
        $\bf{p}_{\mathcal{O}}$ $\gets$ point corresponding to $y$ in $\mathcal{O}$
        $\mathcal{I}_O \gets \bf{p}_{\mathcal{O}}$
    }
}

\caption{Compute Ideal Grasping Region} \label{ideal-region-algorithm}
\end{algorithm}


\subsection{Computing $6$-DOF End-Effector Poses }
\label{sec:6dof}
In order to provide the necessary screw motion to an object using a robotic manipulator, we need to compute a set $\mathcal{E}$ of 6-DOF end-effector poses corresponding to the ideal grasping region, $\mathcal{I}_O$. We use 2 heuristic-based algorithms depending upon our choice of the object--end-effector contact locations, $(\bm{e}^i, \bm{e}^j)$. The dimensions of the bounding box and the gripper width are used to we select two parallel faces $F^i$ and $F^j$ for approximating $\mathcal{I}_B$. The $1^{\text{st}}$ algorithm uses the elements of the grid generated along, $F^i$ or $F^j$, where $y_{\text{avg}}$ $\geq$ $y_{\text{th}}$. Note that only the elements which are occupied by the points of the projected point cloud, $P^i$ are selected. The element centers are used as the contact locations, $(\bm{e}^i, \bm{e}^j)$ and their normals are unit vectors perpendicular to the faces $F^i$ and $F^j$, pointing inwards. The grasp center, $g_c$, which lies halfway along the line joining  $\bm{e}^i$ and $\bm{e}^j$, is computed using the dimensions of the bounding box . The $2^{\text{nd}}$ algorithm randomly select a point, $\bf{p}$ $\in$ $\mathcal{I}_O$, as one of the contact location, $\bm{e}^i$. The orientation of its corresponding normal and dimensions of the bounding box are used to compute the location of the other antipodal contact, $\bm{e}^j$, along with the grasp center, $g_c$. Out of the possible approach directions we select the ones which do not result in collisions between the object and the end-effector. This is done by checking the distance from the grasp center, $g_c$, to the faces of the bounding box corresponding to the approach directions. The gripper geometry is then used to check the feasibility of grasping at $g_c$ from a particular direction as shown in Fig.\ref{Fig: Reference_Directions}. As stated previously for the 2 algorithms, the end-effector orientation is fixed based on the approach direction as shown in Fig \ref{Fig: Reference_Directions} (a).

\section{Experimental Evaluation }

We evaluate the performance of our proposed grasp synthesis approach using (a) simulated partial point clouds (b) data obtained from a RGBD camera and (c) real world experiments with a robot where the objective is to impart a constant screw motion to an object about an associated task screw, $\mathcal{S}$. To evaluate the quality of the grasps, we propose a metric which we call the final grasp evaluation (FGE) metric. Although, our training data has been generated using the task of pivoting an ideal cuboidal object, we show that the proposed neural network-based approach can be used to compute ideal grasping region for objects which are not necessarily cuboidal in shape. 
Thus, our approach is not limited to a particular category, like boxes or bottles, rather it is inherently related to the motion imparted to the object by generating a moment about a particular task screw $\mathcal{S}$. This key point influences our choice of tasks to be used for performing experiments using sensor data.   

\noindent
{\bf Final Grasp Evaluation (FGE):} 
To evaluate the ideal grasping region, $\mathcal{I}_O$, computed using our neural network based approach, we devise a metric to compare it with the output from the algorithm to compute the optimal value of the task-oriented grasp metric in~\cite{fakhari2021computing}. The goal is to check how well our computed regions match with the optimal grasping region that can be computed using~\cite{fakhari2021computing}, but is computationally expensive.
The same point cloud $\mathcal{O}$ and task screw $\mathcal{S}$ is used for both computations by varying the step $5$ of the Algorithm~\ref{ideal-region-algorithm}.
We refer to the 
ideal grasping region computed using~\cite{fakhari2021computing} as $\mathcal{I}^c_O$ and the one approximated using our neural network as $\mathcal{I}^a_O$. Since the metric value computed by our NN is scaled between $0$ and $1$, whereas the output of~\cite{fakhari2021computing} is not scaled, for comparison we need to put them on a common scale. 
We select the top $k$ antipodal contact locations from $\mathcal{I}^a_O$ and top $m$ antipodal contact locations from $\mathcal{I}^c_O$ such that $m>k$. The optimization formulation is used again to evaluate these $(m+k)$ antipodal contact locations by computing their corresponding value of the task-dependent grasp metric, $\eta$. Upon normalizing the values of $\eta$ for the $(m+k)$ contacts, we extract the maximum normalized value, $y_{\text{max}}$, corresponding to the $k$ contact locations from our NN. Higher the value of $y_{\text{max}}$, the closer the antipodal contact locations are to the optimal grasp. 

\begin{figure}[!t]
    \centering
    {\includegraphics[width=0.45\textwidth]{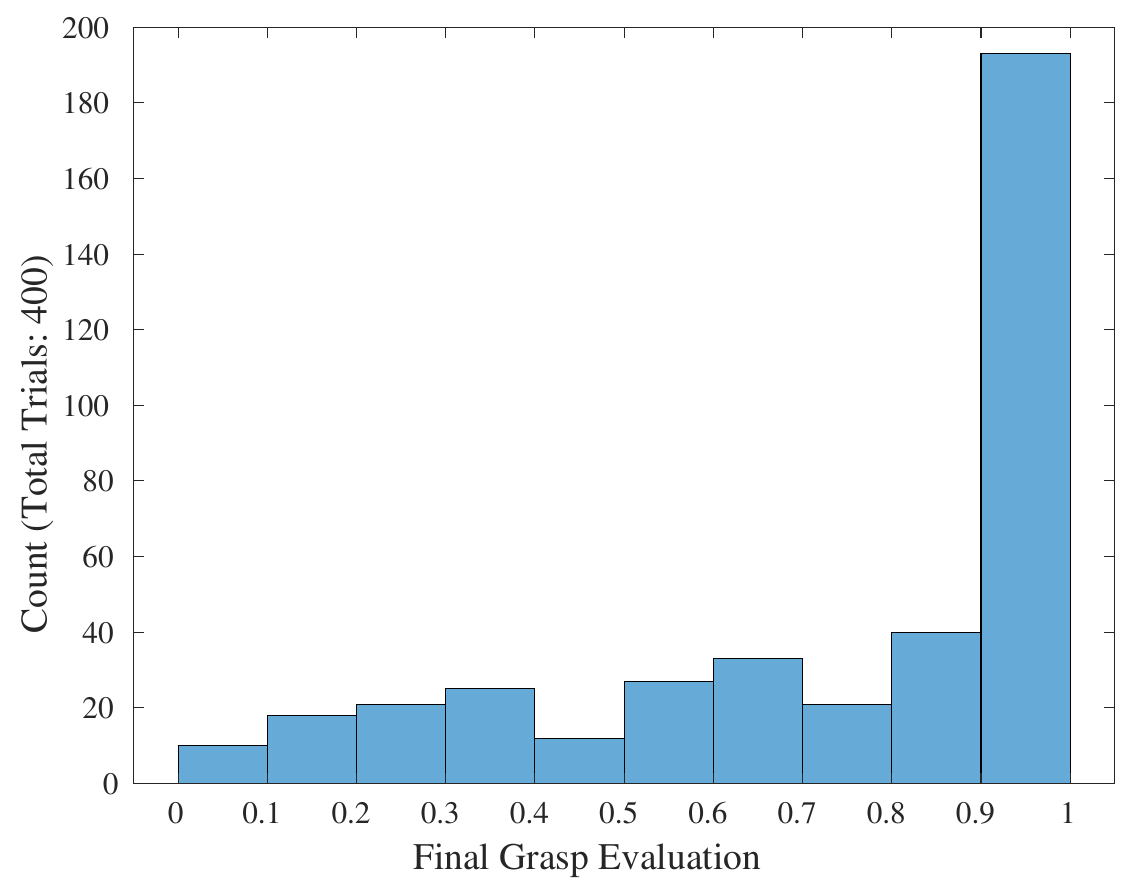}}
    \caption{A histogram showing the distribution of final grasp evaluation, $y_\text{max}$, for each of the 400 trials}
\label{Fig: Simulation_Results}
\end{figure}

\subsection{Evaluating Grasp Synthesis with Simulated Data }

\begin{figure*}
    \centering
    \includegraphics[width=\textwidth]{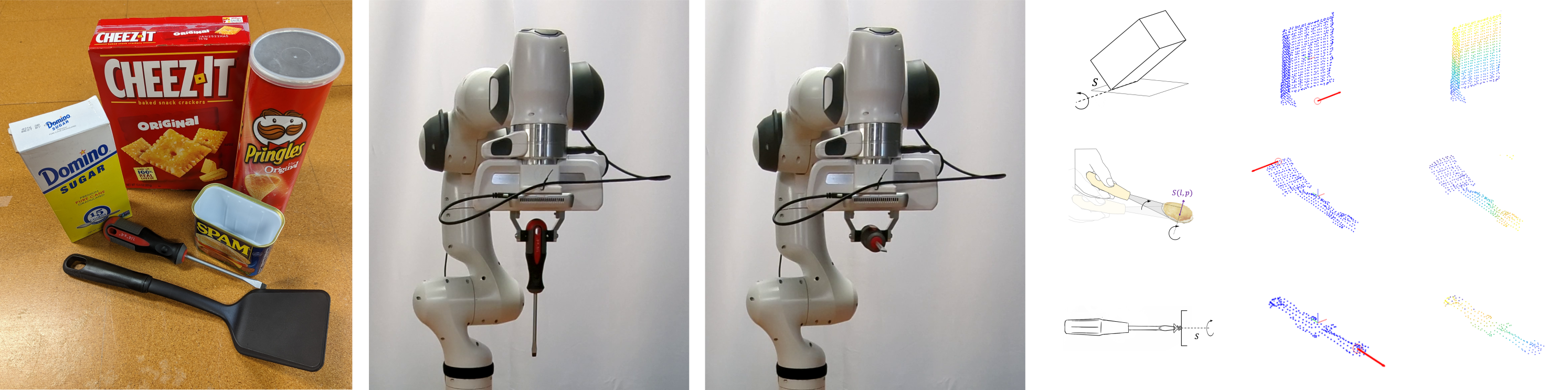}
    \caption{\textbf{\textsc{Experimental Evaluation on Partial Point Clouds:}} The $1^{\text{st}}$ image on the left shows the objects used in our experiments. The $2^{\text{nd}}$ and $3^{\text{rd}}$ images show the $2$ different outcomes for execution trials of the screwing motion. The $4^{\text{th}}$ column of the figure shows the schematic sketches for the task of pivoting, flipping a pancake and operating a screw driver along with their task screws respectively from top to bottom. The $5^{\text{th}}$ column shows the actual partial point clouds for the objects used for performing the same tasks with their associated task screw in red. The $6^{\text{th}}$ column shows the ideal grasping region, as the warmer color, computed using our neural network for the same tasks.}
    \label{fig:my_label}
\end{figure*}

We consider a total of $40$ different objects in our experiments. The CAD and URDF models of the objects are sourced from \cite{liu2021ocrtoc, calli2015ycb} and the PyBullet simulation environment is used to extract the partial point clouds. For each object we 
randomly select $10$ task screws different from the ones in our training data. Thus, we have a total of $400$ trials. 
For each trial, we compute the ideal grasping region using our neural network-based approach and evaluate it using our proposed FGE metric. We set $k=10$ and $m=100$, for computing the FGE metric, $y_\text{max}$. 

\noindent
{\bf Results and Discussion}: Figure~\ref{Fig: Simulation_Results} shows our experimental results displayed as a histogram with the FGE metric $y_\text{max}$ as the $x$-axis. As seen from the figure, $y_\text{max} \geq 0.9$ was observed most number of times. This indicates the ability of our neural network-based approach to compute the ideal grasping region effectively as compared to our optimization formulation. 



\subsection{Evaluating Grasp Synthesis with Real Point Clouds}

Here, we evaluate our approach for computing the ideal grasping region, $\mathcal{I}_O$, on actual sensor data. The data is collected using an Intel Realsense D415 camera mounted on a Franka Emika Panda. Point cloud segmentation is performed using pretrained DeepLab \cite{chen2017rethinking} on RGB images and the corresponding semantic segmentation is used along with the depth channel to get the $3D$ points. We use off-the-shelf algorithms for point cloud processing, normal computation, and computing the bounding box. Vertices of the bounding box are used to compute a fixed object reference frame  $\{ \mathbf{O}\}$, and the bounding box, $\mathcal{B}$, the object point cloud, $\mathcal{O}$ and the task screw, $\mathcal{S}$, are all expressed with respect to $\{ \mathbf{O}\}$. Thus our results do not depend on the absolute pose of the objects in the world frame. The objects used for this experiment are shown in Figure~\ref{fig:my_label}(left). Each object is placed at $3$ different poses, and for each pose, 
 $3$ different camera poses are used giving us a total of $9$ partial point clouds per object. 
 Note that all the objects associated with the tasks above are novel to our grasp synthesis algorithm. However, they are known to the semantic segmentation algorithm of our perception system.


\captionsetup{belowskip=-4pt}
\begin{table}[ht!]
\centering
\begin{tabular}{|c|c|c|}
\cline{1-3}
 \textbf{Object (Task)} & \textbf{Trials} & \textbf{Mean FGE} \\ \cline{1-3}
 Screwdriver (Screwing Motion) & 9 & 0.931 \\ \cline{1-3}
 Spatula (Pivoting) & 9 & 0.928 \\ \cline{1-3}
 CheezIt (Pivoting) & 9 & 0.99 \\ \cline{1-3}
 Domino Sugar (Pivoting) & 9 & 0.784 \\ \cline{1-3}
 Pringles (Pouring) & 9 & 0.967 \\ \cline{1-3}
 Spam Tuna (Pouring) & 9 &  0.835 \\ \cline{1-3}
\end{tabular}
\caption{Experimental Results for Sensor Data}
\label{RGBD Experiments}
\end{table}
\captionsetup{belowskip=-16pt}

We choose three different tasks, namely, screwing motion, pivoting, and pouring as shown in Table~\ref{RGBD Experiments}.
For each task, the task screw, $\mathcal{S}$, is specified. For pivoting, we know the specific edge about which to pivot the CheezIt box and for the screwdriver we know the motion is about an axis along the shank.
However, for any given object pose, it is difficult to know the exact location of the task screw $\mathcal{S}$, and in our implementation we use the bounding box to approximate its location i.e. the point, $\bm{p}$, and its direction, $\bm{l}$. For tasks like pouring, it is difficult to define the task-related constraints imposed on the motion. We leverage an algorithm proposed in~\cite{mahalingam2022human} and extract the task screw from a single kinesthetic demonstration of pouring.
Recall that computing $y_\text{max}$ requires us to solve multiple instances of the optimization formulation~\cite{fakhari2021computing}. To accurately reflect the conditions of the task, we need to take into consideration any object-environment contacts if necessary. This is being done for the task of pivoting with all the $3$ objects. In this set of experiments, for each trial we set $k = 5$ and $m = 50$ while computing FGE. 

\noindent
{\bf Results and Discussion:}
The results of the experiments are displayed in table \ref{RGBD Experiments}. The first column specifies the object and its associated task and the second column shows the number of trials conducted for each object. The third column shows the mean value of the final grasp evaluation, $y_\text{max}$, for each of the 9 trials. The mean values of FGE for all the objects are closer to $1$ which is desirable. This shows that we can sample antipodal contact locations, $(\bm{e}^i, \bm{e}^j)$, from the ideal grasping region that can generate the desired magnitude of moment about the task screw even for partial point clouds obtained from a RGBD camera.

\subsection{Task Oriented Grasping with a Robot}
We evaluate our method using three different tasks (1) operating a screwdriver (2) pivoting a box about its edge and (3) pouring using the same hardware setup as above, where the robot grasps the objects and performs the motion (please see companion video). 

\captionsetup{belowskip=-8pt}
\begin{table}[ht!]
\centering
\begin{tabular}{|c|c|c|c|}
\cline{1-4}
\textbf{Object (Task)} & \textbf{Trials}  &  \textbf{Successful} & \textbf{Mean FGE} \\ \cline{1-4}
 Pringles (Pouring) & 5 & 5 & 0.91  \\ \cline{1-4}
 CheezIt (Pivoting) & 8 & 6 & 0.99 \\ \cline{1-4}
 Domino Sugar (Pivoting) & 8 & 7 & 0.96  \\ \cline{1-4}
 Spam Tuna (Pouring) & 6 & 5 & 0.892 \\ \cline{1-4}
  Screwdriver & 15 & 5 & 0.891 \\ \cline{1-4}
\end{tabular}
\caption{Experimental Results for Grasping and Motion Planning}
\label{Grasping_with_Motion_Planning}
\end{table}
\captionsetup{belowskip=-16pt}

Similar to the previous set of experiments, we assume that the task screws are known beforehand. 
After computing the ideal grasping region, $\mathcal{I}_O$, we use the $1^{\text{st}}$ algorithm in Section~\ref{sec:6dof} to compute the corresponding set of end-effector poses, $\mathcal{E}$, and randomly select one end-effector pose $\mathbf{E}$ for grasping. For the pivoting and screwing motion, apart from computing the ideal grasping region, $\mathcal{I}_O$, their corresponding task screw to compute the final pose using Screw linear interpolation (ScLERP)~\cite{fakhari2021motion}. A ScLERP-based motion planner~\cite{sarker2020screw} is then used to generate a motion plan from the initial end-effector pose. For the task of pouring, the sequence of constant screw segments extracted from a kinesthetic demonstration is used along with the ScLERP-based motion planner to generate motion plans~\cite{mahalingam2022human}. Note that we use position control for executing the motion and set a fixed upper limit for the maximum force that can be applied by the grippers. The grasp computation takes approximately $3$ seconds on an Intel $i7$ processor with $32$ GB RAM and all the experiments are done with real-time grasp computation and planning.

\noindent
{\bf Results and Discussion: } The results are shown in Table \ref{Grasping_with_Motion_Planning}. The $2^{\text{nd}}$ column shows the total number of trials conducted for each task and the $3^{\text{rd}}$ column shows the number of successful trials. The last column shows the mean value of the final grasp evaluation metric for all the trials. The mean value of FGE is close to $1$ indicating that upon sampling contact locations from the ideal grasping region, $\mathcal{I}_O$, we can get good grasps. But this does not necessarily translate to successful task execution as we use a different criteria for evaluating the success for each task. For example, in the case of a screwdriver, keeping the orientation constant throughout the motion after grasping, is desirable. Although the screwdriver was grasped and picked up for every trial, its orientation was not maintained at all times (see Figure~\ref{fig:my_label}, 2nd and 3rd panels from left). We classified a trial as failure if the screwdriver was not horizontal after picking up. The key reason for the failures is that the screwdriver has a slightly curved end and we are not controlling the gripper force directly, which makes it difficult for the parallel jaw grippers to maintain the contact force to prevent the rotational slip. 

For pivoting, in $13$ trials (out of $16$), the robot was able to successfully pivot the objects
The $3$ failures were due to error in computing the task screw from the partial point cloud data thereby introducing inaccuracies in the motion plan. This causes the box to slip at the object-environment contact or it is pushed against the surface triggering the robot's collision detection mechanism. For pouring, there was $10$ successful trials and $1$ failures. The one failure was due to failure in grasp due to slipping.

\section{Conclusion and Future Work: }
\label{sec:concl}
In this paper we present a novel grasp synthesis approach for task-oriented grasping from partial point cloud data of objects, where we want to give a motion to the objects after grasping. Many high level tasks like pivoting, pouring, and turning a handle are tasks where the goal of manipulation after grasping is to give a particular motion to the object. Our approach of exploiting the screw geometry of motion to define a task allows us to integrate the grasp synthesis algorithm with a class of ScLERP based motion planning approaches~\cite{sarker2020screw}. Furthermore, since we use a very simple geometric representation of the partial point cloud as a cuboid to compute the grasping region on the object, we do not need any prior knowledge about the objects to generate the grasps. We present both simulation and experimental results to show the validity of our approach for realistic manipulation tasks with real sensor data. 

Our approach is complementary to the approaches of task-oriented grasping where human knowledge is used to label the good grasping regions. In the future we plan to combine labels on good grasping regions along with the bounding boxes on partial point clouds of the objects to incorporate human intuition and knowledge within our framework. This will enable the robot to grasp at handles of cups or places meant to hold objects to which our current approach is agnostic. This will also potentially allow the robot to exploit human knowledge about tasks as well as the geometric structure of motion that constrains manipulation, and is difficult to obtain from human labeling. 

\bibliographystyle{IEEEtran}
\bibliography{references}

\begin{thebibliography}{10}
\providecommand{\url}[1]{#1}
\csname url@samestyle\endcsname
\providecommand{\newblock}{\relax}
\providecommand{\bibinfo}[2]{#2}
\providecommand{\BIBentrySTDinterwordspacing}{\spaceskip=0pt\relax}
\providecommand{\BIBentryALTinterwordstretchfactor}{4}
\providecommand{\BIBentryALTinterwordspacing}{\spaceskip=\fontdimen2\font plus
\BIBentryALTinterwordstretchfactor\fontdimen3\font minus
  \fontdimen4\font\relax}
\providecommand{\BIBforeignlanguage}[2]{{%
\expandafter\ifx\csname l@#1\endcsname\relax
\typeout{** WARNING: IEEEtran.bst: No hyphenation pattern has been}%
\typeout{** loaded for the language `#1'. Using the pattern for}%
\typeout{** the default language instead.}%
\else
\language=\csname l@#1\endcsname
\fi
#2}}
\providecommand{\BIBdecl}{\relax}
\BIBdecl

\bibitem{kokic2020learning}
M.~Kokic, D.~Kragic, and J.~Bohg, ``Learning task-oriented grasping from human
  activity datasets,'' \emph{IEEE Robotics and Automation Letters}, vol.~5,
  no.~2, pp. 3352--3359, 2020.

\bibitem{wen2022catgrasp}
B.~Wen, W.~Lian, K.~Bekris, and S.~Schaal, ``Catgrasp: Learning category-level
  task-relevant grasping in clutter from simulation,'' in \emph{2022
  International Conference on Robotics and Automation (ICRA)}.\hskip 1em plus
  0.5em minus 0.4em\relax IEEE, 2022, pp. 6401--6408.

\bibitem{fakhari2021computing}
A.~Fakhari, A.~Patankar, J.~Xie, and N.~Chakraborty, ``Computing a
  task-dependent grasp metric using second-order cone programs,'' in \emph{2021
  IEEE/RSJ International Conference on Intelligent Robots and Systems
  (IROS)}.\hskip 1em plus 0.5em minus 0.4em\relax IEEE, pp. 4009--4016.

\bibitem{Borst2004}
C.~{Borst}, M.~{Fischer}, and G.~{Hirzinger}, ``Grasp planning: how to choose a
  suitable task wrench space,'' in \emph{IEEE International Conference on
  Robotics and Automation (ICRA)}, vol.~1, 2004, pp. 319--325.

\bibitem{pollard1994parallel}
N.~S. {Pollard}, ``Synthesizing grasps from generalized prototypes,'' in
  \emph{IEEE International Conference on Robotics and Automation (ICRA)},
  vol.~3, 1996, pp. 2124--2130.

\bibitem{haschke2005task}
R.~Haschke, J.~J. Steil, I.~Steuwer, and H.~Ritter, ``Task-oriented quality
  measures for dextrous grasping,'' in \emph{International Symposium on
  Computational Intelligence in Robotics and Automation}.\hskip 1em plus 0.5em
  minus 0.4em\relax IEEE, 2005, pp. 689--694.

\bibitem{krug2016analytic}
R.~Krug, A.~J. Lilienthal, D.~Kragic, and Y.~Bekiroglu, ``Analytic grasp
  success prediction with tactile feedback,'' in \emph{IEEE International
  Conference on Robotics and Automation (ICRA)}, 2016, pp. 165--171.

\bibitem{song2020grasping}
S.~Song, A.~Zeng, J.~Lee, and T.~Funkhouser, ``Grasping in the wild: Learning
  6dof closed-loop grasping from low-cost demonstrations,'' \emph{IEEE Robotics
  and Automation Letters}, vol.~5, no.~3, pp. 4978--4985, 2020.

\bibitem{mahalingam2022human}
D.~Mahalingam and N.~Chakraborty, ``Human-guided planning for complex
  manipulation tasks using the screw geometry of motion,'' in \emph{2023
  International Conference on Robotics and Automation (ICRA)}, 2023.

\bibitem{sarker2020screw}
A.~Sarker, A.~Sinha, and N.~Chakraborty, ``On screw linear interpolation for
  point-to-point path planning,'' in \emph{2020 IEEE/RSJ International
  Conference on Intelligent Robots and Systems (IROS)}.\hskip 1em plus 0.5em
  minus 0.4em\relax IEEE, 2020, pp. 9480--9487.

\bibitem{patankar2020hand}
A.~Patankar, A.~Fakhari, and N.~Chakraborty, ``Hand-object contact force
  synthesis for manipulating objects by exploiting environment,'' in \emph{2020
  IEEE/RSJ International Conference on Intelligent Robots and Systems
  (IROS)}.\hskip 1em plus 0.5em minus 0.4em\relax IEEE, 2020, pp. 9182--9189.

\bibitem{newbury2022deep}
R.~Newbury, M.~Gu, L.~Chumbley, A.~Mousavian, C.~Eppner, J.~Leitner, J.~Bohg,
  A.~Morales, T.~Asfour, D.~Kragic \emph{et~al.}, ``Deep learning approaches to
  grasp synthesis: A review,'' \emph{arXiv preprint arXiv:2207.02556}, 2022.

\bibitem{kokic2017affordance}
M.~Kokic, J.~A. Stork, J.~A. Haustein, and D.~Kragic, ``Affordance detection
  for task-specific grasping using deep learning,'' in \emph{2017 IEEE-RAS 17th
  International Conference on Humanoid Robotics (Humanoids)}.\hskip 1em plus
  0.5em minus 0.4em\relax IEEE, 2017, pp. 91--98.

\bibitem{detry2017task}
R.~Detry, J.~Papon, and L.~Matthies, ``Task-oriented grasping with semantic and
  geometric scene understanding,'' in \emph{2017 IEEE/RSJ International
  Conference on Intelligent Robots and Systems (IROS)}.\hskip 1em plus 0.5em
  minus 0.4em\relax IEEE, 2017, pp. 3266--3273.

\bibitem{do2018affordancenet}
T.-T. Do, A.~Nguyen, and I.~Reid, ``Affordancenet: An end-to-end deep learning
  approach for object affordance detection,'' in \emph{2018 IEEE international
  conference on robotics and automation (ICRA)}.\hskip 1em plus 0.5em minus
  0.4em\relax IEEE, 2018, pp. 5882--5889.

\bibitem{ardon2020self}
P.~Ard{\'o}n, E.~Pairet, Y.~Petillot, R.~P. Petrick, S.~Ramamoorthy, and K.~S.
  Lohan, ``Self-assessment of grasp affordance transfer,'' in \emph{2020
  IEEE/RSJ International Conference on Intelligent Robots and Systems
  (IROS)}.\hskip 1em plus 0.5em minus 0.4em\relax IEEE, 2020, pp. 9385--9392.

\bibitem{deng20213d}
S.~Deng, X.~Xu, C.~Wu, K.~Chen, and K.~Jia, ``3d affordancenet: A benchmark for
  visual object affordance understanding,'' in \emph{Proceedings of the
  IEEE/CVF Conference on Computer Vision and Pattern Recognition}, 2021, pp.
  1778--1787.

\bibitem{fang2020learning}
K.~Fang, Y.~Zhu, A.~Garg, A.~Kurenkov, V.~Mehta, L.~Fei-Fei, and S.~Savarese,
  ``Learning task-oriented grasping for tool manipulation from simulated
  self-supervision,'' \emph{The International Journal of Robotics Research},
  vol.~39, no. 2-3, pp. 202--216, 2020.

\bibitem{murali2021same}
A.~Murali, W.~Liu, K.~Marino, S.~Chernova, and A.~Gupta, ``Same object,
  different grasps: Data and semantic knowledge for task-oriented grasping,''
  in \emph{Conference on Robot Learning}.\hskip 1em plus 0.5em minus
  0.4em\relax PMLR, 2021, pp. 1540--1557.

\bibitem{sun2021gater}
M.~Sun and Y.~Gao, ``Gater: Learning grasp-action-target embeddings and
  relations for task-specific grasping,'' \emph{IEEE Robotics and Automation
  Letters}, vol.~7, no.~1, pp. 618--625, 2021.

\bibitem{ferrari1992planning}
C.~Ferrari and J.~F. Canny, ``Planning optimal grasps.'' in \emph{ICRA},
  vol.~3, no.~4, 1992, p.~6.

\bibitem{mahler2017dex}
J.~Mahler, J.~Liang, S.~Niyaz, M.~Laskey, R.~Doan, X.~Liu, J.~A. Ojea, and
  K.~Goldberg, ``Dex-net 2.0: Deep learning to plan robust grasps with
  synthetic point clouds and analytic grasp metrics,'' \emph{In: Proceedings of
  Robotics: Science and Systems (RSS)}, 2017.

\bibitem{ten2017grasp}
A.~ten Pas, M.~Gualtieri, K.~Saenko, and R.~Platt, ``Grasp pose detection in
  point clouds,'' \emph{The International Journal of Robotics Research},
  vol.~36, no. 13-14, pp. 1455--1473, 2017.

\bibitem{alliegro2022end}
A.~Alliegro, M.~Rudorfer, F.~Frattin, A.~Leonardis, and T.~Tommasi,
  ``End-to-end learning to grasp via sampling from object point clouds,''
  \emph{IEEE Robotics and Automation Letters}, vol.~7, no.~4, pp. 9865--9872,
  2022.

\bibitem{lou2021collision}
X.~Lou, Y.~Yang, and C.~Choi, ``Collision-aware target-driven object grasping
  in constrained environments,'' in \emph{2021 IEEE International Conference on
  Robotics and Automation (ICRA)}.\hskip 1em plus 0.5em minus 0.4em\relax IEEE,
  2021, pp. 6364--6370.

\bibitem{lou2020learning}
------, ``Learning to generate 6-dof grasp poses with reachability awareness,''
  in \emph{2020 IEEE International Conference on Robotics and Automation
  (ICRA)}.\hskip 1em plus 0.5em minus 0.4em\relax IEEE, 2020, pp. 1532--1538.

\bibitem{kasaei2021mvgrasp}
H.~Kasaei and M.~Kasaei, ``Mvgrasp: Real-time multi-view 3d object grasping in
  highly cluttered environments,'' \emph{arXiv preprint arXiv:2103.10997},
  2021.

\bibitem{mousavian20196}
A.~Mousavian, C.~Eppner, and D.~Fox, ``6-dof graspnet: Variational grasp
  generation for object manipulation,'' in \emph{Proceedings of the IEEE/CVF
  International Conference on Computer Vision}, 2019, pp. 2901--2910.

\bibitem{grant2014cvx}
M.~Grant and S.~Boyd, ``Cvx: Matlab software for disciplined convex
  programming, version 2.1,'' 2014.

\bibitem{he2016deep}
K.~He, X.~Zhang, S.~Ren, and J.~Sun, ``Deep residual learning for image
  recognition,'' in \emph{Proceedings of the IEEE conference on computer vision
  and pattern recognition}, 2016, pp. 770--778.

\bibitem{ioffe2015batch}
S.~Ioffe and C.~Szegedy, ``Batch normalization: Accelerating deep network
  training by reducing internal covariate shift,'' in \emph{International
  conference on machine learning}.\hskip 1em plus 0.5em minus 0.4em\relax pmlr,
  2015, pp. 448--456.

\bibitem{ba2016layer}
J.~L. Ba, J.~R. Kiros, and G.~E. Hinton, ``Layer normalization,'' \emph{arXiv
  preprint arXiv:1607.06450}, 2016.

\bibitem{liu2021ocrtoc}
Z.~Liu, W.~Liu, Y.~Qin, F.~Xiang, M.~Gou, S.~Xin, M.~A. Roa, B.~Calli, H.~Su,
  Y.~Sun, and P.~Tan, ``Ocrtoc: A cloud-based competition and benchmark for
  robotic grasping and manipulation,'' 2021.

\bibitem{calli2015ycb}
B.~Calli, A.~Singh, A.~Walsman, S.~Srinivasa, P.~Abbeel, and A.~M. Dollar,
  ``The ycb object and model set: Towards common benchmarks for manipulation
  research,'' in \emph{2015 international conference on advanced robotics
  (ICAR)}.\hskip 1em plus 0.5em minus 0.4em\relax IEEE, 2015, pp. 510--517.

\bibitem{chen2017rethinking}
L.-C. Chen, G.~Papandreou, F.~Schroff, and H.~Adam, ``Rethinking atrous
  convolution for semantic image segmentation,'' \emph{arXiv preprint
  arXiv:1706.05587}, 2017.

\bibitem{fakhari2021motion}
A.~Fakhari, A.~Patankar, and N.~Chakraborty, ``Motion and force planning for
  manipulating heavy objects by pivoting,'' in \emph{2021 IEEE/RSJ
  International Conference on Intelligent Robots and Systems (IROS)}.\hskip 1em
  plus 0.5em minus 0.4em\relax IEEE, 2021, pp. 9393--9400.

\end{thebibliography}

\end{document}